\documentclass[runningheads]{llncs}
\usepackage[T1]{fontenc}
\usepackage{graphicx, amsmath, xcolor, siunitx, cleveref, wrapfig, subcaption}
\usepackage{cite}

\usepackage{booktabs}
\usepackage{amsfonts}

\def\method{\emph{STORK}}

\begin{document}
\title{{STORK}: Spatio-Temporal Observation of\\uterine contRactions via neural networKs}

%
\titlerunning{{STORK}}
%
 \author{Melissa Schween\inst{1,2,3} \and
 Tristan Gottwald\inst{2,3} \and
 Jordina Aviles Verdera\inst{1,2,3} \and
 Lisa Story\inst{4} \and
 Mary Rutherford\inst{4} \and
 Jana Hutter\inst{1,2,3}}

\authorrunning{M. Schween et al.}
%
 \institute
     {CAIMED, Leibniz University Hannover, Hannover, Germany 
 \and
     L3S, Leibniz University Hannover, Hannover, Germany 
 \and
     Institute of Information Processing, Leibniz University Hannover, Germany 
 \and
     Department for Biomedical Engineering, King's College London, London, UK
     \email{schween@tnt.uni-hannover.de}
 }

\maketitle              
\begin{abstract}

Uterine contractions in fetal MRI are typically identified manually and discarded, limiting insights into contraction dynamics. We formalize Uterine Contractile Activity Detection (UCAD) as a weakly-supervised learning problem and introduce \method{}, a multi-instance learning model trained on dynamic MRI series using only coarse, series-level labels. \method{} factorizes 3D spatio-temporal convolutions into parallel branches across temporal hyperplanes to capture coherent tissue motion without the cost of full 4D convolutions. Per-frame embeddings, combining intensity and Demons-estimated displacement fields, are aggregated by a linear mean-pooling head. This ensures that frame-level contraction scores can be recovered post-hoc without frame-level training supervision. Evaluated on around 700 multi-vendor dynamic fetal MRI series, \method{} achieves a series-level AUROC of 95.0\% and AUPRC of 94.6\%, substantially outperforming 3D ResNet and ConvNeXt baselines. Grad-CAM analysis suggests that the model draws on predictive features extending beyond the placenta into the uterine tissue, offering an automated tool for richer phenotyping of uterine behavior.

\keywords{Fetal MRI  \and Relaxometry \and Multi Instance Learning.}

\end{abstract}
\section{Introduction}

Frequent antenatal health assessments are conducted typically by Ultrasound (US) to assess growth and Cardiotocography (CTG) to quantify uterine contractile activity, fetal state and reaction to contractions. Both, however, struggle to provide a comprehensive assessment. US is operator-dependent and suffers from a limited field-of-view particularly at late gestational age. CTG lacks spatial resolution and functional insights into the placental core. Dynamic Magnetic Resonance Imaging (MRI) addresses some of these challenges, providing whole-uterus spatio-temporal coverage and diverse functional contrasts. However, while used diagnostically in CTG, uterine contractions hamper MR imaging and contribute to artifacts. Currently, affected images are manually identified and discarded. Automated contraction detection would allow for explicit quantification of contractile activity, data-driven rejection, and adaptive imaging protocols.

\begin{figure}[tb]
  \centering
  \includegraphics[width=0.95\textwidth]{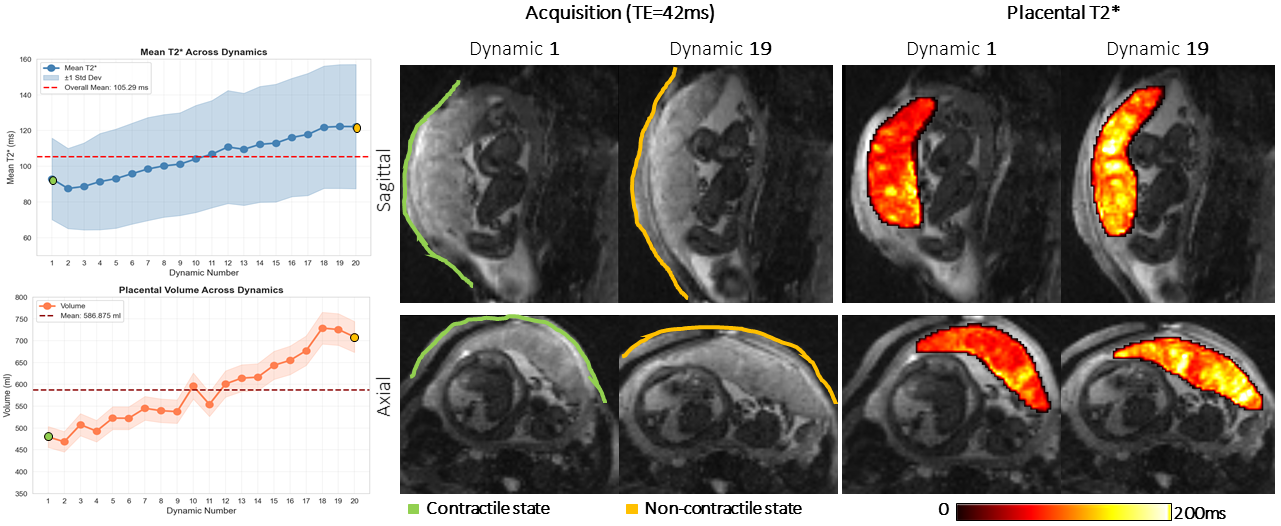}
   \caption{Illustration of contractile activity captured during a dynamic Multi-Echo Gradient Echo MRI scan, showing quantitative mean T2* and volume (left), acquired data at the first echo time (center) and quantitative T2* maps (right).}
   \label{fig:contraction-intro}
\end{figure}

\Cref{fig:contraction-intro} illustrates the quantitative T2* and volumetric changes associated with contractions, motivating their detection from dynamic MRI data.
We introduce a weakly-supervised solution trained on series-level labels, called \method. Our contributions include (1) a formal problem definition of \textit{Uterine Contractile Activity Detection}, (2) a hyperplane-factorized MIL architecture for dynamic MRI under weak series-level supervision, (3) interpretability analysis by leveraging a linear mean-pooling backbone, aggregating per-frame embeddings into series-level predictions while ensuring that frame-level scores can be recovered post-hoc, (4) discovery of discriminative features beyond the placenta, and (5) comprehensive comparison against convolutional and temporal baselines.

\section{Related Work}
\label{sec:related_work}

\subsection{Clinical Background and Uterine MRI}
\label{subsec:clinical_background}
Fetal dynamic MRI, particularly T2* relaxometry, is increasingly used for contractile assessment, capturing reductions in placental T2*, volume changes and qualitative patterns such as darkened bands through the placental core \cite{AvilesVerdera2025con,AbaciTurk2022,Hutter2022,Sinding2016,Dellschaft2020}. Prior work has used image-based features to differentiate contraction types~\cite{Dellschaft2020}, manually identified contractions to exclude corrupted quantitative data~\cite{AvilesVerderaRAT2025, slator2019}, and demonstrated downstream hypoxic effects in the fetal brain and liver. Most recently, a retrospective multi-vendor analysis introduced a step towards automatic classification using thresholding on placental T2* curves~\cite{AvilesVerdera2025con}.

\subsection{Machine Learning Background}
\label{subsec:ml_background}
Prior machine learning applications to fetal MRI focused on anatomical segmentation and quality enhancement~\cite{meshaka2023artificial, hai2025applications}, typically treating motion as an artifact. In contrast, we utilize motion, via displacement fields, as the primary discriminative signal for contractile activity. For sequence-level annotations, Multiple Instance Learning (MIL) treats a sequence (e.g., frames) as a bag of instances, labeled positive if at least one instance is positive~\cite{maron1997framework, ilse2018attention}. Permutation-invariant pooling aggregates per-instance features into a bag representation~\cite{ilse2018attention}. Learned attention also yields instance-level scores, enabling weakly-supervised localization without frame annotations. This paradigm was successfully adopted from computational pathology~\cite{campanella2019clinical, lu2021data}, to lung cancer diagnosis~\cite{chen2022lung}. For per-instance spatio-temporal encoding, 3D CNN backbones like ResNet~\cite{he2016deep} or ConvNeXt~\cite{liu2022convnet} remain a high-performing choice in medical imaging~\cite{roy2025mednext}. Recent architectures leverage tensor factorization for high-dimensional data like MRI volumes~\cite{zhou2025spatiotemporal, vishnevskiy2020deep}. For instance, FlowVN~\cite{vishnevskiy2020deep} replaces full 4D kernels with orthogonal 3D convolutions across dimension subsets to capture directional spatio-temporal correlations within its regularization term while reducing complexity.
While FlowVN uses this factorization for rapid 4D flow reconstruction, we repurpose it as a feed-forward feature extractor. Instead of using all four 3D combinations, \method{} only uses the three combinations including the temporal components as one of the dimensions, since the purely spatial (xyz) plane contributes no motion information relevant to contractile activity.
Similar factorization techniques have also been adopted outside medical imaging to represent dynamic 3D scenes by accelerating radiance fields~\cite{Cao_2023_CVPR, fridovich2023k}.
\section{\textit{Uterine Contractile Activity Detection} (UCAD)}\label{sec:problem_def}
Uterine contraction is essentially muscular activity of the myometrium manifesting as coordinated tissue deformation over space and time with the effects varying depending on the tissue and proximity of the placenta. Detection is complicated by other moving structures (maternal respiration, bowel peristalsis, fetal motion) producing local change of similar magnitude, and by placental-related pathologies manifesting with similar structural changes. We therefore approach detection through observable induced changes rather than by explicitly defining contractions.

\noindent Let a dynamic MRI acquisition be a temporal sequence of 3D volumes
\begin{equation} \label{eq:aquired_data}
    \mathbf{\hat{X}} = (\mathbf{\hat{x}}_1, \mathbf{\hat{x}}_2, \ldots, \mathbf{\hat{x}}_T), \quad
    \mathbf{\hat{x}}_t \in \mathbb{C}^{E \times \hat{H} \times \hat{W} \times \hat{D}},
\end{equation}
where $T$ is the number of acquired time frames, $\hat{H} \times \hat{W} \times \hat{D}$ is the spatial extent, and $E$ the number of acquired echo times with each image consisting of magnitude and phase. We work in image space; raw k-space is outside the scope.

\noindent Let $\mathbf{r} \in \Omega \subset \mathbb{R}^3$ denote spatial position, and let $\delta_t(\mathbf{r}) \in \mathbb{R}_{\geq 0}$ denote a local measure of change at location $\mathbf{r}$ between time $t-1$ and $t$, derived from $\mathbf{x}_{t-1}$ and $\mathbf{x}_t$. Conceptually, the observed change superimposes contractile motion $\delta_{t,\text{con}}$ and non-contractile motion $\delta_{t,\text{noise}}$, such that
$\delta_t(\mathbf{r}) = \delta_{t,\text{con}}(\mathbf{r}) \oplus \delta_{t,\text{noise}}(\mathbf{r})$,
where $\oplus$ denotes superposition of motion sources rather than a separable additive decomposition. The two components are not individually observable, and isolating $\delta_{t,\text{con}}$ is itself part of the learning problem. A contraction corresponds to the existence of at least one sub-region $\mathcal{R}_{i,t} \subset \Omega$ in which the contractile component is sustained above a threshold $\tau$ subject to spatial and temporal coherence conditions: 
\begin{equation} \label{eq:contraction_region}
    \delta_{t,\text{con}}(\mathbf{r}) > \tau \quad
    \forall\, \mathbf{r} \in \mathcal{R}_{i,t}.
\end{equation}

Spatially, $\mathcal{R}_{i,t}$ must be a connected region of volume at least $a_{\min}$, ruling out isolated voxels and scattered responses caused by noise. Temporally, the activity must persist across a minimum duration, holding over all frames in a window $[t, t+\Delta]$ with $\Delta \geq \Delta_{\min}$. \Cref{eq:contraction_region} is a schema rather than an operational definition as no closed form is known for $\delta_{t,\text{con}}, \tau, a_{\min}$, or $\Delta_{\min}$. These quantities, together with the rejection of non-contractile movement, are what the model must learn.
\noindent The learning objective is a mapping $f$ from the sequence of MRI volumes $\mathbf{\hat{X}}$ to a series-level contraction probability, indicating whether a contraction occurs at any time and anywhere within the series:
\begin{equation} \label{eq:objective}
    f : \mathbb{C}^{T \times E \times \hat{H} \times \hat{W} \times \hat{D}} \to [0,1],
    \qquad \hat{y} = f(\mathbf{\hat{X}}) \approx p(y = 1 \mid \mathbf{\hat{X}}),
\end{equation}
where $y \in \{0,1\}$ is the series-level label. Beyond this series-level decision, the quantity of ultimate interest is \emph{where} and \emph{when} a contraction occurs (the sub-regions $\mathcal{R}_{i,t}$ of \cref{eq:contraction_region}), which we treat as a secondary objective, to be recovered from the series-level model rather than dedicated supervision.

\section{Methodology}\label{sec:methodology}
\begin{figure}[t]
  \centering
  \includegraphics[width=1.0\textwidth]{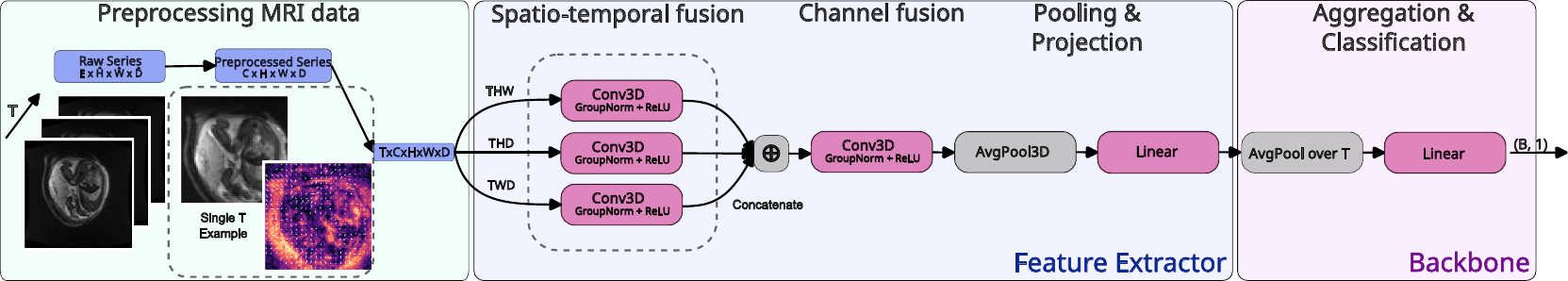}
   \caption{\method{} network architecture composed of a \textit{preprocessing stage} (green), a \textit{feature extractor} (blue) and \textit{backbone} (pink). Preprocessing stage transforms each series into a 4-channel input combining magnitude images with Demons displacement fields. The feature extractor processes the 4D volume by leveraging 3D convolutions on hyperplanes always involving the temporal dimension. The backbone aggregates information over the time dimension by average pooling combined with a linear projection.}
   \label{fig:model-pipe}
\end{figure}

We instantiate the mapping $f$ of \cref{eq:objective} as a three-stage pipeline, consisting of a preprocessing step, a feature extractor and a backbone (see \cref{fig:model-pipe}).

\subsection{Preprocessing}\label{subsec:preprocessing}
Each series is reduced from the acquisition $\mathbf{\hat{X}}$ to the model input $\mathbf{X}$ through \textit{cropping}, \textit{per-voxel displacement}, \textit{normalization}, and \textit{resampling}, resulting in
\begin{equation} \label{eq:model_input}
    \mathbf{X} = (\mathbf{x}_1, \mathbf{x}_2, \ldots, \mathbf{x}_T), \quad
    \mathbf{x}_t \in \mathbb{R}^{C \times H \times W \times D},
\end{equation}
where $H \times W \times D$ is the fixed spatial target resolution and  $C=4$ the channel dimension. A dilated binary segmentation mask of the placenta~\cite{Hall2024} defines the cropping and normalization regions but is not applied to the network input.
Frames are \textit{cropped} to the axis-aligned bounding box of the placenta mask, retaining broader uterine context. For each pair of consecutive frames, we estimate a \textit{per-voxel displacement} $\mathbf{u}_t(\mathbf{r}) \in \mathbb{R}^3$ by Demons registration~\cite{demons1998} and clip it to a physiologically plausible range. This provides an explicit representation of inter-frame tissue motion. Stacking the single-echo intensity with the displacement components yields the input $\mathbf{x}_t = [\,\text{intensity}, \mathbf{u}_t(\mathbf{r})\,]$ of \cref{eq:model_input}. The single-echo choice is justified in \Cref{sec:ablation}.
Channels are \textit{normalized} per series to preserve temporal intensity changes that signal a contraction. Using the dilated placenta mask as the region interest, the intensity channel is \textit{normalized} using a single mean and standard deviation computed across all $T$ frames. Displacement components are standardized similarly. Finally, frames are resized via trilinear interpolation to a fixed spatial resolution. \textit{Resampling} is performed last to ensure displacement field calculation and normalization operate at native resolution, giving a series tensor of shape $(T, C, H, W, D)$.

\subsection{Network Architecture}\label{subsec:network_architecture}
The network architecture comprises a \textit{feature extractor} and a \textit{backbone}, mapping the spatio-temporal input of shape $(T, C, H, W, D)$ to a series-level classification.

\paragraph{Feature Extractor.}
A contraction is defined jointly over space and time (\cref{eq:contraction_region}), so a feature extractor that encodes each frame in isolation cannot represent it.
Rather than applying a full 4D convolution, we factorize it into three parallel 3D convolutional branches with kernel size $k=9$, each operating on a different spatio-temporal hyperplane ($THW$, $THD$, $TWD$), similar to approaches in \cite{vishnevskiy2020deep, Cao_2023_CVPR, fridovich2023k}. This preserves the spatio-temporal structure, while avoiding the cost of a full 4D convolution.
We remove the spatial hyperplane ($HWD$), since the spatio-temporal structure captured by the other hyperplanes is more relevant for classification than the volume's spatial structure alone.
Coherent tissue motion over time is encoded directly rather than reconstructed from independent frame features. Each branch outputs $d$ feature maps which are concatenated and fused back to $d$ channels via a depth-wise convolution, mixing the space-time evidence at each location.  Finally, global 3D average pooling over the spatial dimensions reduces the tensor to a temporal sequence of vectors, and a linear layer projects each frame to an embedding $\mathbf{z}_t \in \mathbb{R}^{d_{\text{emb}}}$ that encodes its contextual neighbors.

\paragraph{Backbone.}
To aggregate the sequence $\mathbf{z} = (\mathbf{z}_1, \ldots, \mathbf{z}_T)$ under weak series-level labels, we use a parameter-free temporal mean and a linear layer:
\begin{equation} \label{eq:meanpool}
    \hat{y} = \sigma\!\left( \mathbf{w}^\top \, \frac{1}{T} \sum_{t=1}^{T} \mathbf{z}_t + b \right),
\end{equation}
where $\sigma$ is the sigmoid function. Because the mean and the linear mapping commute, the series logit decomposes exactly into independent per-frame contributions ($\mathbf{w}^\top \mathbf{z}_t + b$). This linearity enables frame-level localization post-hoc without explicit frame annotations. The model is trained end-to-end minimizing the binary cross-entropy between $\hat{y}$ and the series label $y$. To counter class imbalance, positive instances are weighted by the ratio of negative-to-positive series in the training set.

\section{Experiments}\label{sec:experiments}
We evaluate performance using \textit{Area Under the Receiver Operating Characteristic} (AUROC) and \textit{Area Under the Precision-Recall curve} (AUPRC), with the latter providing a more stringent evaluation under the class imbalance present in our dataset (see \Cref{subsec:dataset}). Series-level metrics assess if a series is correctly classified as containing contractile activity via the aggregated series logit. Frame-level metrics instead evaluate standalone frame predictions against frame-level label, assessing localized contractile activity within a series, serving as a secondary, finer-grained view of model behavior. All results report the mean and standard deviation of ten random runs.

\subsection{Dataset}
\label{subsec:dataset}
Dynamic MRI data was acquired in pregnant participants as part of ethically approved studies (21/LO/0742, 22/YH/0210, 23/LO/0685, 21/SS/0082, 19/LO/ 0852) on clinical MRI scanners (0.55T, 1.5T, 3T). A multi-echo gradient echo planar imaging sequence with imaging parameters varying by field strength was employed \cite{AvilesVerdera2025con}.  A total of 757 data sets were available, acquired in 549 pregnant participants (mean gestational age=32.2 weeks). Placental masks were obtained with a previously trained network~\cite{Hall2024}, T2* fitting was performed via mono-exponential least squares fit. Only cases defined contractile by both  proposed thresholding mean T2* \cite{AvilesVerdera2025con} and by gamma-fitting the placental mean T2* value were maintained (57 cases dropped). Only around 40\% of the cases got labeled as showing contractions, leading to an imbalanced dataset. Note that this labeling detects contractile activity through placental T2* signal only, whereas our model operates on the image crop including surrounding anatomy (see \Cref{subsec:preprocessing}). Frame- and series-level predictions may therefore respond to contractile activity outside the placenta, not captured by the weak labels. Data is partitioned at the \emph{subject} level to prevent leakage. Splits are stratified by (cohort, label) to balance the positive rate per cohort. Subjects with manual frame-level labels are reserved for the test set. The remaining split is 70\%, 10\%, 20\% (train, val, test).

\subsection{Implementation details}\label{subsec:implementation}
Frames are resized by trilinear interpolation to $H \times W \times D = 64 \times 64 \times 32$. Segmentation masks are dilated by 10\,mm before cropping, and the crop bounding box is padded by 15 voxels to retain surrounding anatomical context.
The intensity channel uses echo 1 of the multi-echo acquisition, following the single echo approach ablated in \Cref{sec:ablation}. Frame-to-frame displacement fields are computed with SimpleITK Demons registration (20 iterations, standard deviation 1.0), clipped to $\pm 20$ mm, and z-scored independently per channel using mask-region statistics; intensity is z-scored analogously.
No data augmentation is applied. Parameters are optimized with AdamW~\cite{loshchilov2019decoupled}, learning rate $5 \times 10^{-4}$, weight decay $10^{-2}$, under a schedule that warms up linearly for 3 epochs before decaying by cosine annealing. Gradients are clipped to a norm of 10 for stability.
Models are trained with batch size 1 for up to 40 epochs, selecting the checkpoint with the best series-level validation AUROC and stopping early after 10 epochs without improvement. Each configuration is trained with 10 random seeds, which jointly control the subject-level data split and model initialization; we report mean $ \pm $ std across seeds.
All reported experiments were conducted on a workstation with a Nvidia RTX 4090 with 24\,GB VRAM and a Intel(R) Core(TM) i9-11900K.

\subsection{Main Results}\label{subsec:results}
We introduce a naive baseline from components commonly used for image~\cite{khan2020survey, xu2023resnet, hasanah2023deep} and time-series~\cite{he2019temporal, farha2019ms} processing. A ResNet-10~\cite{he2016deep} extracts spatial features from each volume independently, and a Temporal Convolutional Network (TCN)~\cite{lea2016temporal} aggregates the resulting embeddings across time. We further include a second baseline replacing the ResNet backbone with ConvNeXt~\cite{liu2022convnet, woo2023convnext}, incorporating design choices from Vision Transformers to achieve comparable performance while maintaining a convolutional nature. Such architectures had recently strong performance in medical imaging tasks~\cite{ roy2025mednext}.
As a third point of comparison, we replace the mean-pooling aggregation of \method{} with attention-based MIL pooling~\cite{ilse2018attention}, the standard learned aggregation in weakly-supervised settings. Since attention pooling is permutation-invariant and therefore blind to acquisition order, we add a fixed sinusoidal positional encoding~\cite{vaswani2017attention} to the frame embeddings before the attention module, making temporal position available to the attention scores.

Comparing \method{} to the baselines in \Cref{tab:main_results} shows that the ResNet baseline achieves a performance just above random, while the ConvNeXt baseline performs slightly better. As both TCN-based baselines lack a per-frame head, we cannot provide frame-wise metrics. 
The variant of \method{} using attention retains a per-frame linear head and is therefore evaluated at both levels. It recovers most of the performance but remains below \method{} at the series level and exhibits higher seed variance at the frame level. Learned attention thus offers no advantage over the parameter-free temporal mean here, while being less stable under weak supervision.
Our model achieves strong series-level performance, with an AUROC of 95.0\% and AUPRC of 94.6\%. Notably, \method{} also performs well on frame-level metrics despite receiving no frame-level supervision during training.
Furthermore, \method{}'s architecture allows for the processing of 70 temporal slices per second with a a peak VRAM usage of 3.4\,GB making our approach suitable for near real-time processing at moderate hardware requirements. 
Overall, \method{} clearly outperforms the TCN-based baselines by a wide margin and exceeds the attention variant at the series level, while, unlike the baselines, providing frame-level localization despite training only on series-level labels.

\begin{table}[t]
    \centering
    \caption{Test-set results over seeds. Best values in \textbf{bold}.}
    \label{tab:main_results}
    \resizebox{\linewidth}{!}{
    \begin{tabular}{l|cccc}
        \toprule
        & \multicolumn{4}{c}{Metrics (mean $\pm$ std)} \\
        \cmidrule(lr){2-5}
        Approach & Series AUROC $\uparrow$ & Series AUPRC $\uparrow$ & Frame AUROC $\uparrow$ & Frame AUPRC $\uparrow$ \\
        \midrule
        \method{} (ours) & \textbf{0.950} $\pm$ 0.011 & \textbf{0.946} $\pm$ 0.016 & \textbf{0.797} $\pm$ 0.015 & \textbf{0.674} $\pm$ 0.026 \\
        \method{} $+$ Attention & 0.915 $\pm$ 0.022 & 0.899 $\pm$ 0.036 & 0.763 $\pm$ 0.057 & 0.657 $\pm$ 0.091 \\
        ResNet Baseline & 0.556 $\pm$ 0.096 & 0.520 $\pm$ 0.096 & n/a & n/a\\
        ConvNeXt Baseline & 0.610 $\pm$ 0.064 & 0.558 $\pm$ 0.073 & n/a & n/a \\
        \bottomrule
    \end{tabular}
    }
\end{table}

\subsection{Latent Space Interpretability}\label{subsec:latent_space}
\begin{figure}[tbp]
    \centering
    \begin{subfigure}[b]{0.4\textwidth}
        \centering
        \includegraphics[width=\textwidth]{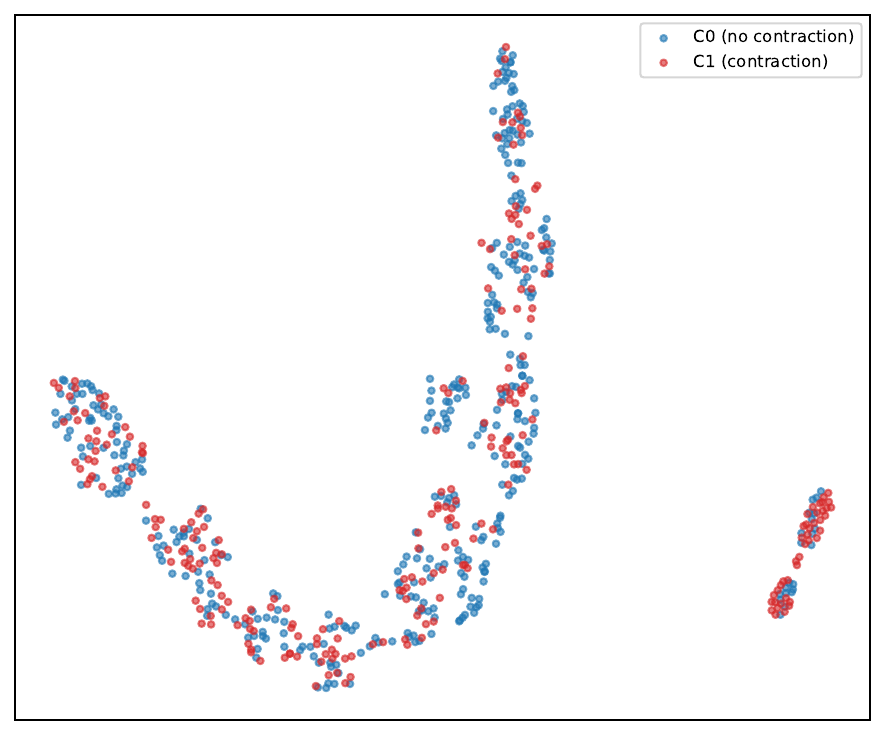}
        \caption{Baseline latent space}
        \label{fig:latent_space_baseline}
    \end{subfigure}
    \hfill 
    \begin{subfigure}[b]{0.4\textwidth}
        \centering
        \includegraphics[width=\textwidth]{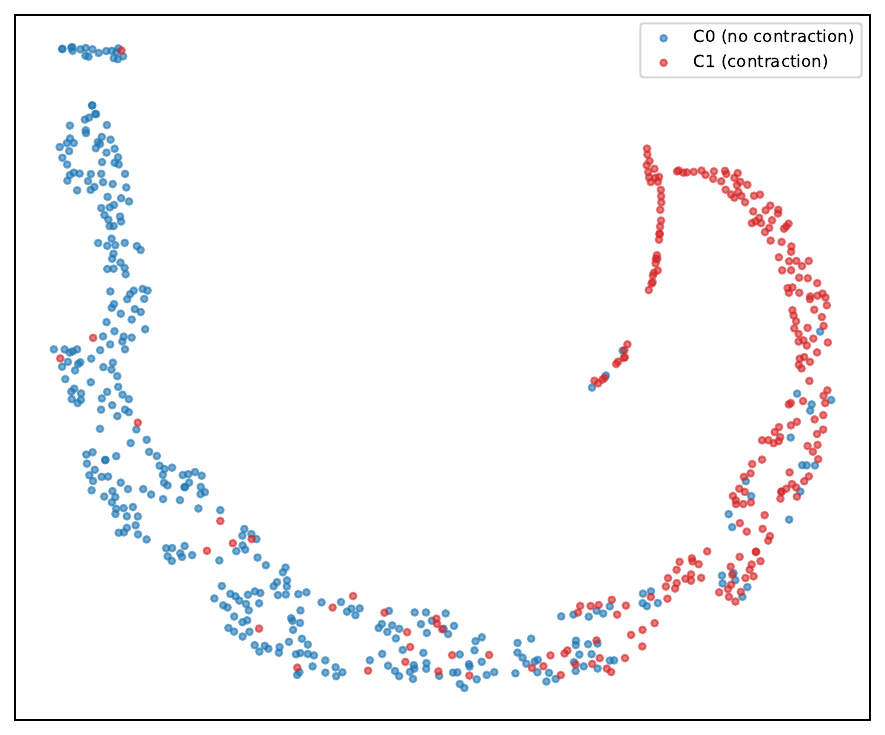}
        \caption{\method{} latent space}
        \label{fig:latent_space_dachs}
    \end{subfigure}
    \caption{t-SNE plot of the latent spaces for ResNet baseline (a) and \method{}(b), showing separation of non-contractile (blue) and contractile (red) cases.}
    \label{fig:latent_space}
\end{figure}

The feature extractor maps high-dimensional MRI volumes into a low-dimensional latent space. \Cref{fig:latent_space} visualizes these representations for the test set using t-SNE~\cite{van2008visualizing} to project sequence embeddings into two dimensions, comparing the ResNet baseline (\cref{fig:latent_space_baseline}) against our \method{} architecture (\cref{fig:latent_space_dachs}). While the ResNet baseline fails to separate contraction (red) and non-contraction (blue) sequences, \method{} achieves a distinct separation.
This highlights the strength of our model, learning a meaningful representation of the data, while the baseline struggles to learn distinguishing features.

\subsection{Ablations} \label{sec:ablation}
\begin{figure}[t]
  \centering
  \includegraphics[width=0.8\textwidth]{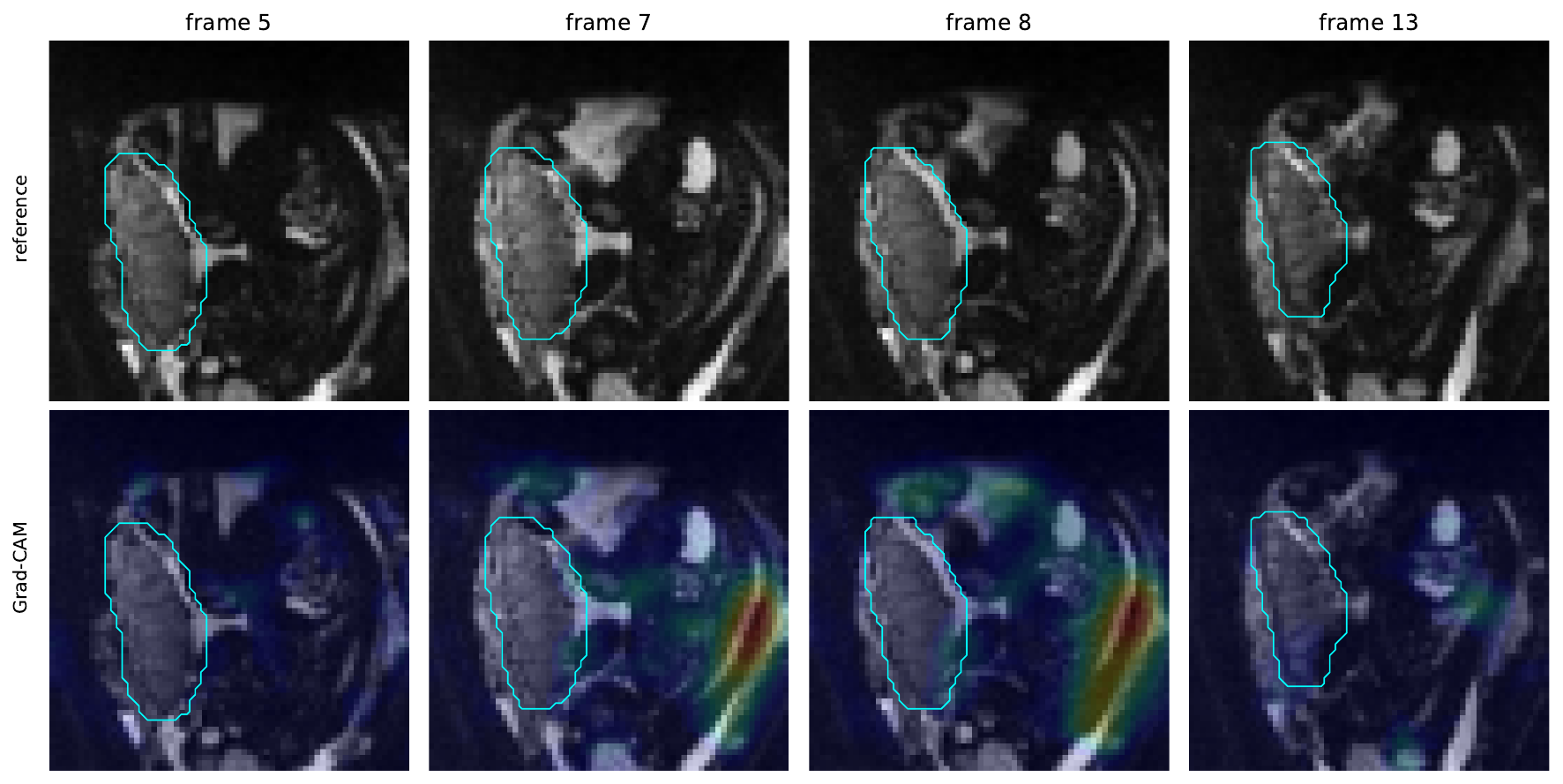}
   \caption{Frame-wise Grad-CAM. Top: reference intensity with placenta contour (cyan); bottom: same frames with the Grad-CAM overlay for the series logit.}
   \label{fig:grad_cam}
\end{figure}

We ablate the choice of echo to isolate its contribution, and use Grad-CAM to inspect where the model's predictions originate. \Cref{tab:echo_ablation} evaluates the influence of the acquired echoes. Echo~1 (indexing from 0) attains the best value on two metrics. With marginal differences between single and multi-echo, we adopt Echo~1 for \method{} to reduce computational cost.
Frame-level scores are obtained post-hoc from the per-frame decomposition of the series logit (\cref{eq:meanpool}) and evaluated against the manually annotated test frames.
Frame-wise analysis with Grad-CAM~\cite{selvaraju2017gradcam} showed (\cref{fig:grad_cam}) that the model often attends to areas outside of the placenta, here the uterine wall. This is consistent with our hypothesis that other parts of the uterus carry relevant signal. These attributions offer temporally resolved localization that may support broader clinical application, but remain qualitative and cannot be validated against the placenta-derived labels. A quantitative attribution analysis is left to future work.

\begin{table}[t]
    \centering
    \caption{Echo ablation comparing single and the best combined echoes at the series and frame level. Best values in \textbf{bold}. Echoes are indexed from 0.}
    \label{tab:echo_ablation}
    \resizebox{\linewidth}{!}{
    \begin{tabular}{l|cccc}
        \toprule
        & \multicolumn{4}{c}{Metrics (mean $\pm$ std)} \\
        \cmidrule(lr){2-5}
        Preprocessing & Series AUROC $\uparrow$ & Series AUPRC $\uparrow$ & Frame AUROC $\uparrow$ & Frame AUPRC $\uparrow$\\
        \midrule
        Echo 0 & 0.937 $\pm$ 0.014 & 0.934 $\pm$ 0.017 & 0.784 $\pm$ 0.025 & 0.654 $\pm$ 0.041 \\
        Echo 1 & \textbf{0.950} $\pm$ 0.011 & 0.946 $\pm$ 0.016 & 0.797 $\pm$ 0.015 & \textbf{0.674} $\pm$ 0.026 \\
        Echo 2 & \textbf{0.950} $\pm$ 0.011 & 0.943 $\pm$ 0.020 & 0.795 $\pm$ 0.016 & 0.658 $\pm$ 0.024 \\
        Echo 0, 1, 2 & 0.948 $\pm$ 0.010 & \textbf{0.948} $\pm$ 0.015 & \textbf{0.799} $\pm$ 0.019 & 0.664 $\pm$ 0.043 \\
    \bottomrule
    \end{tabular}
    }
\end{table}

\section{Conclusion and Discussion}
\method{} introduces an end-to-end, learning-based method for detecting uterine contractile activity in dynamic MRI. Unlike prior work constrained by hand-crafted pipelines on mean placental T2* curves~\cite{AvilesVerdera2025con}, our approach learns discriminative features directly from the image, extending across the uterus rather than the placenta alone. Leveraging a large dataset of over 700 dynamic series, \method{} attains strong series-level performance under weak supervision, and recovers frame-level localization without frame-level training. A central limitation is the training label quality, which relies on automated T2* curve proxies. The introduced label noise may result in the model learning signal from regions the proxy labels never verify. Despite this, automated, non-invasive contraction monitoring provides a richer phenotype of uterine behavior. Clinically, it offers insights into complications like fetal growth restriction and may help predict how a fetus tolerates the physiological stress of labor. Methodologically, our framework for recovering frame-level localization from weak labels extends naturally to other transient biomedical events. Finally, integrating this model for real-time detection during acquisition could allow scanning protocols to dynamically adapt temporal resolution or extend acquisitions during live events.

\begin{credits}
This work was supported by the Bavarian State Ministry of Health, Care and Prevention [project EndoKI], DFG Heisenberg [502024488], ERC StG EARTHWORM [101165242], ERC Proof-of-concept grant SYNCWORM [101293293] and CAIMed - Lower Saxony Center for Artificial Intelligence and Causal Methods in Medicine [ZN4257] funding. The authors have no competing interests to declare.
\end{credits}
%
%
\bibliographystyle{splncs04}
\bibliography{references}
\end{document}